\documentclass[runningheads]{llncs}
\usepackage{graphicx}
\usepackage{amsfonts}
\usepackage{multirow}
\usepackage[colorlinks=true,citecolor=blue,urlcolor=blue]{hyperref}

\begin{document}
\title{DGM-DR: Domain Generalization with Mutual Information Regularized Diabetic Retinopathy Classification}
\titlerunning{DGM-DR}
%

\author{Aleksandr Matsun\inst{*} \and
Dana O. Mohamed\inst{*} \and
Sharon Chokuwa\inst{*} \and
Muhammad Ridzuan \and
Mohammad Yaqub} 


\authorrunning{A. Matsun et al.}
\def\thefootnote{*}\footnotetext{These authors contributed equally to this work. 

Dana O. Mohamed is the corresponding author.}\def\thefootnote{\arabic{footnote}}

\institute{Mohamed Bin Zayed University of Artificial Intelligence, Abu Dhabi, UAE
\email{\{aleksandr.matsun,dana.mohamed,sharon.chokuwa,muhammad.ridzuan,
mohammad.yaqub\}@mbzuai.ac.ae}}

\maketitle              
\begin{abstract}
The domain shift between training and testing data presents a significant challenge for training generalizable deep learning models. As a consequence, the performance of models trained with the independent and identically distributed (i.i.d) assumption deteriorates when deployed in the real world. This problem is exacerbated in the medical imaging context due to variations in data acquisition across clinical centers, medical apparatus, and patients. Domain generalization (DG) aims to address this problem by learning a model that generalizes well to any unseen target domain. Many domain generalization techniques were unsuccessful in learning domain-invariant representations due to the large domain shift. Furthermore, multiple tasks in medical imaging are not yet extensively studied in existing literature when it comes to DG point of view. In this paper, we introduce a DG method that re-establishes the model objective function as a maximization of mutual information with a large pretrained model to the medical imaging field. We re-visit the problem of DG in Diabetic Retinopathy (DR) classification to establish a clear benchmark with a correct model selection strategy and to achieve robust domain-invariant representation for an improved generalization. Moreover, we conduct extensive experiments on public datasets to show that our proposed method consistently outperforms the previous state-of-the-art by a margin of 5.25\% in average accuracy and a lower standard deviation. Source code available at \url{https://github.com/BioMedIA-MBZUAI/DGM-DR}.

\keywords{Domain Generalization  \and Diabetic Retinopathy \and Mutual Information Regularization}
\end{abstract}
\section{Introduction}
Medical imaging has become an indispensable tool in diagnosis, treatment planning, and prognosis. Coupled with the introduction of deep learning, medical imaging has witnessed tremendous progress in recent years. Notwithstanding, a major challenge in the medical imaging field is the domain shift problem, where the performance of a trained model deteriorates when for instance tested on a dataset that was acquired from a different device or patient population than the original dataset. This problem is especially prominent in tasks, where acquiring large-scale annotated datasets from one center is costly and time-consuming. Domain generalization (DG) \cite{DG_Survey} aims to alleviate this challenge by training models that can generalize well to new unseen domains, without the need for extensive domain-specific data collection and annotation.

DG in medical image analysis still requires extensive research, however there already exist a handful of works examining it. One of those works includes utilizing an adversarial domain synthesizer to create artificial domains using only one source domain to improve the generalizability of the model in downstream tasks \cite{RW_adv_consis}. Although such method can synthesize a wide range of possible domains, it usually suffers from the ability to mimic realistic domain shifts. Another method is applying test-time augmentations such that the target image resembles the source domain, thus reducing the domain shift and improving generalization \cite{RW_DG_Histo_Cell}. Moreover, DRGen \cite{drgen} combines Fishr \cite{Fishr} and Stochastic Weight Averaging Densely (SWAD) \cite{SWAD} to achieve domain generalization in Diabetic Retinopathy (DR) classification. In DRGen, Fishr \cite{Fishr} is used to make the model more robust to variations in the data by penalizing large differences in the gradient variances between in-distribution and out-of-distribution data, and SWAD \cite{SWAD} is used to seek flatter minima in the loss landscape of the model. DRGen is currently state-of-the-art in DR classification, however it has been evaluated using samples from the testing set which makes it harder to assess its true generalizability.

In natural images, the domain generalization problem has been explored extensively compared to medical imaging analysis. Some of the DG methods proposed over the past ten years include domain alignment \cite{intro_domain_align_1}, meta-learning \cite{intro_meta-learning}, style transfer \cite{intro_style_transfer}, and regularization methods \cite{intro_DG_regularization}. More recently, the authors of \cite{MIRO} utilize a large pretrained model to guide a target model towards generalized feature representation through mutual information regularization. Another DG regularization method that can be applied orthogonally to many DG algorithms is SWAD \cite{SWAD}, which improves domain generalizability by seeking flat minima in the loss landscape of the model. The flatter minima indicate that the loss is not changing significantly in any direction, thus reducing the risk of the model overfitting to domain biases  \cite{SWAD}. However, when adapting a DG approach that demonstrates a good performance on natural images, there is no guarantee of a similar performance on medical imaging applications due to the typical complex nature of such problems.

DR is a complication of Diabetes Mellitus that affects the eyes and can lead to vision loss or blindness. It is caused by damage to the blood vessels in the retina due to high blood sugar levels, which often leads to blood leakage onto the retina \cite{medical_DR_paper_1_General}. This can cause swelling and distortion of vision. The prevalence of DR is increasing worldwide due to the growing number of people with diabetes. However, early detection and management of DR is critical to the prevention of vision deterioration or loss. DR can be classified into 4 classes: mild, moderate, severe, and proliferative 
. Some of the visible features that are used to classify the first 3 classes include microaneurysms, retinal hemorrhages, intraretinal microvascular abnormalities (IRMA), and venous caliber changes, while pathologic preretinal neovascularization is used to classify proliferative DR \cite{medical_DR_paper_3_visual_features}.

In this paper, we propose DGM-DR, a Domain Generalization with Mutual information regularized Diabetic Retinopathy classifier. Our main contributions are as follows:
\begin{itemize}
        \item We introduce a DG method that utilizes mutual information regularization with a large pretrained oracle model.
        \item We show the improvement of our proposed solution on the DR classification task over the previous state-of-the-art in both performance and robustness through rigorous investigations.
        \item We set a clear benchmark with the correct DG model selection method inline with standard DG protocols for the task of DR classification.
\end{itemize}

\begin{figure}[t]
\includegraphics[width=\textwidth]{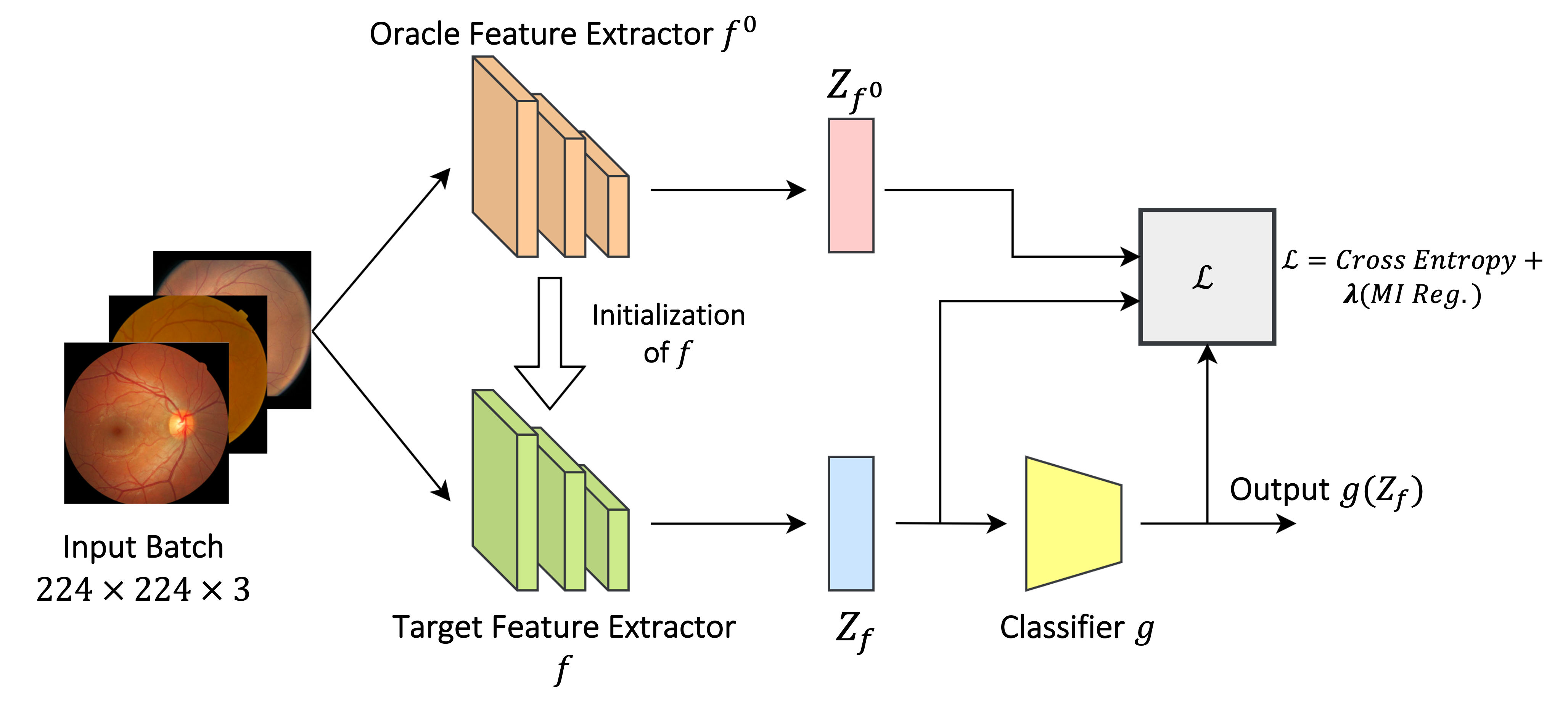}
\caption{Overview of the proposed method, DGM-DR. It consists of the oracle $f^0$ and target $f$ feature extractor, where $f$ is initialized by the weights of $f^0$. For each sampled mini-batch, feature representations are extracted using both feature extractors $f^0$ and $f$. Features $Z_f$ are then passed to classifier $g$. Lastly, the loss - a linear combination of cross entropy and mutual information regularization loss - is calculated, and $f$ and $g$ are updated.} \label{Figure:model}
\end{figure}
\section{Methodology}
Our work is inspired by \cite{MIRO}, which aims to improve model generalizability when classifying natural images. In DGM-DR, we re-establish the domain generalization objective as a maximization of mutual information with a large pretrained model, named the oracle, to address DR classification. We aim to make the distribution of feature representations of the target model close to the generalized one of the oracle by maximizing the mutual information between both. The oracle model is trained on a large-scale diverse dataset that contains information on many different domains in order to approximate it as closely as possible to a true oracle, which is a model that can generalize to any domain and is inaccessible in practice. Figure \ref{Figure:model} shows an overview of DGM-DR's process. Initially, the oracle's weights are used to initialize the target model's feature extractor. Then, for each mini-batch, the oracle feature extractor $f^0$ and the target feature extractor $f$ are used to extract feature representations $Z_{f^0}$ and $Z_f$, respectively. The features $Z_f$ are passed to the classifier $g$ to produce the output. The oracle model is chosen as ImageNet pretrained ResNet-50 \cite{resnet50} for a realistic and fair comparison with other DG algorithms. It is shown in  ~\cite{barber2004algorithm} that maximization of the lower bound of the mutual information between $Z_{f^0}$ and $Z_f$ is equivalent to minization of the term \ref{eq:min_term}

\begin{equation}
\label{eq:min_term}
 \mathbb{E} _{Z_{f^{0}},Z_f} \bigl[ log |\Sigma(Z_{f})| + \|Z_{f^0} - \mu(Z_f) \|_{\Sigma(Z_{f})^{-1}}^2
\end{equation}

The final loss is calculated using Equation \ref{eq:MIRO}:

\begin{equation}
\label{eq:MIRO}
\mathcal{L}(h) = \mathcal{E} _{S}(h) + \lambda \mathbb{E} _{Z_{f^{0}},Z_f} \bigl[ log |\Sigma(Z_{f})| + \|Z_{f^0} - \mu(Z_f) \|_{\Sigma(Z_{f})^{-1}}^2 \bigr]
\end{equation}
where $\mathcal{E} _{S} (.) = \sum _{d=1}^{m} \mathcal{E} _{S_{d}}(.)$ is an empirical loss over $m$ source domains, which was chosen as cross-entropy loss, $\lambda$ is the regularization coefficient, and $\| x \|_{A} = \sqrt{x^TAx}$. The model $h$ is modeled as a composition of a feature extractor $f$ and a classifier $g$, hence $h = f \circ g$. Finally, the variational distribution that approximates the oracle model is modeled as a Gaussian distribution with mean vector $\mu(Z_f)$ and covariance matrix $\Sigma(Z_{f})$. $\|Z_{f^0} - \mu(Z_f) \|$ enforces the mean feature representation $Z_f$ to be as close as possible to the oracle feature representation $Z_{f^0}$ when the variance term  $\Sigma(Z_{f})$ is low \cite{MIRO}. We anticipate that this optimization will yield robust representations, despite the substantial distribution shift between the oracle pretrained on natural images and the finetuning task involving retinal images. This is based on our hypothesis regarding the oracle's generalizability to any domain, owing to its extensive, diverse, and semantically rich features that surpass those found in any other medical dataset. The regularization term $\lambda$ aims to minimize the variance in the target features and encourage similarity between the oracle and target features. This, in turn, facilitates the learning of domain-invariant representations that generalize well across different domains.

\section{Experimental Setup}
\subsection{Datasets}
\begin{figure}[t]
\includegraphics[width=\textwidth]{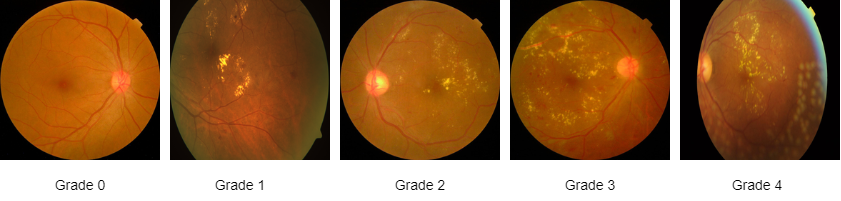}
\caption{Sample images from different DR classes obtained from APTOS \cite{APTOS}.} \label{Figure:DR-images}
\end{figure}

We utilize the four datasets used by \cite{drgen}, which are EyePACS \cite{EyePACS}, APTOS \cite{APTOS}, Messidor and Messidor-2 \cite{MESSIDOR}. The 4 datasets are composed of 5 classes of 5 grades from 0 to 4: No DR (Grade 0), mild DR (Grade 1), moderate DR (Grade 2), severe DR (Grade 3), and proliferative DR (Grade 4). These datasets were acquired from various geographical regions, encompassing India, America, and France \cite{APTOS,EyePACS,MESSIDOR}. As a result, domain shift emerges, due to the variations in the employed cameras \cite{EyePACS,atwany2022deep}, and the difference in population groups. Figure \ref{Figure:DR-images} shows example images for the 5 DR classes. A breakdown of the distribution of the classes is given in Table \ref{table:datasets}. In all 4 datasets, there is a high imbalance between the no DR class and the other 4 DR classes. 

\subsection{Data Augmentations}
All fundus images are resized to $224 \times 224 \times 3$. We perform histogram equalization with a probability $p=0.5$, horizontal flip and color jitter by a value of 0.3 in brightness, contrast, saturation, and hue with $p=0.3$.

\subsection{Evaluation Methods} 
We utilize the DomainBed \cite{domainbed} evaluation protocols for fair comparison with DRGen \cite{drgen} and other DG algorithms. The appropriate DG model selection method used is the training-domain validation set following DomainBed \cite{domainbed}, in which we split each training domain into training and validation subsets, pool the validation subsets together to create an overall validation set, and finally choose the model that maximizes the accuracy on the overall validation set. We use 20\% of the source training data for validation. We evaluate the performance scores using leave-one-domain-out cross validation, and average the cases where a specific domain is used as a target domain and the others as source domains.

We also perform comparisons of the proposed and existing DG approaches with the Empirical Risk Minimization (ERM) technique that aims to minimize in-domain errors. Interestingly, \cite{domainbed} argues that carefully training a model using ERM achieves a near state-of-the-art performance. This was tested on a range of baselines and was shown to outperform a few DG models.

\subsection{Implementation Details}
We implement all our models using the PyTorch v1.7 framework. The experiments were run on 24GB Quadro RTX 6000 GPU. The backbone used is ResNet-50 pretrained on ImageNet. We use the Adam optimizer \cite{adam-optimizer} with a learning rate of $5e-5$ and no weight decay, chosen experimentally. The model was trained in 5000 steps. The batch size was fixed to 32 images. The $\lambda$ regularization coefficient was set to 1.0. Different values of lambda were experimented with, and the results are given in \ref{table:lambda-table}.

To compare against other DG methods, we reproduce the results of all algorithms using the same implementation details mentioned previously for a fair comparison. For Fishr \cite{Fishr}, we set the Fishr lambda ($\lambda$) to 1000, penalty anneal iteration ($\gamma$) of 1500 and an exponential moving average of 0.95. For DRGen \cite{drgen}, we use SWAD as the model selection method as opposed to the test-domain validation used in the original paper \cite{drgen}, which is not suitable for DG evaluation. Moreover, we use the data augmentations in the official implementations of Fishr \cite{Fishr} and DRGen \cite{drgen} for the respective algorithms, otherwise we use DGM-DR's augmentations. Finally, we use SWAD as the model selection method when combining DGM-DR with the SWAD \cite{SWAD} algorithm.

\section{Results}
Table \ref{table:main-results} compares the performance of DGM-DR with three other methods, including the previous state-of-the-art DRGen. The experiments for the main results were repeated three times using three different seeds, and the average accuracy and standard deviation across the runs are reported. DGM-DR achieves $>5\%$ increase in the average accuracy when compared with the other DG methods (Fishr and DRGen) and 1\% increase compared to ERM-based model \cite{ERM}.
\begin{table}[t]
    \scriptsize
    \centering
    \caption{Multi-class classification results with ERM and DG methods averaged over three runs. The best accuracy (\%) is highlighted in bold.}
    \begin{tabular}{l|cccc|c}
        \hline
        \textbf{ Algorithm }     & \textbf{ APTOS } & \textbf{ EyePACS } & \textbf{ Messidor } & \textbf{ Messidor-2 } & \textbf{ Average Accuracy } \\ \hline
        \textbf{ERM}\cite{ERM}           & 62.83       & 73.01        & \textbf{66.88}         & 65.26            & 66.99$\pm$4.3                   \\
        \textbf{Fishr}\cite{Fishr}        & 56.49        & 68.24          & 61.53           & 62.11            & 62.09 $\pm$4.8                   \\
        \textbf{DRGen}\cite{drgen}        & 54.53        & \textbf{73.87}          & 52.03           & 69.13             & 62.39$\pm$9.3                   \\ \hline
        \textbf{DGM-DR}        & \textbf{65.39}        & 70.12         & 65.63           & \textbf{69.41}             & 67.64$\pm$2.4                   \\ 
        \textbf{DGM-DR + SWAD}\cite{SWAD} & 65.15        & 71.92          & 65.66           & 68.96             & \textbf{67.92$\pm$3.2}                   \\ \hline
    \end{tabular}
    \label{table:main-results}
\end{table}

\subsection{Ablation Studies}

\begin{table}[t]
    \scriptsize
    \centering
    \caption{Results of changing the oracle pretraining datasets, methods, and backbones. SWAG* is using a batch size of 16, while the rest of the experiments are using a batch size of 32. The average accuracy and the standard deviation across the 4 domains in a single run are given, with the best accuracy (\%) highlighted in bold.}
    \begin{tabular}{l|l|cccc|c}
    \hline
    \textbf{Dataset}                                        & \textbf{Pre-training} & \textbf{APTOS} & \textbf{EyePACS} & \textbf{Messidor} & \textbf{Messidor 2} & \multicolumn{1}{c}{\textbf{Average Accuracy}} \\ \hline
    \multicolumn{1}{c|}{\multirow{3}{*}{\textbf{ImageNet}}} & ERM                & \textbf{65.39}              & 70.12                & 65.63                 & \textbf{69.41}                   & \textbf{67.64$\pm$2.5}                                              \\ \cline{2-7} 
    \multicolumn{1}{c|}{}                                   & Barlow Twins        & 60.66          & 73.45            & 55.57             & 61.18               & 62.71$\pm$7.6                                          \\ \cline{2-7} 
    \multicolumn{1}{c|}{}                                   & MoCo v3             & 56.90          & 72.69            & 65.77             & 68.41               & 65.94$\pm$6.7                                          \\ \hline
    \multirow{2}{*}{\textbf{CLIP}}                          & CLIP \tiny{(ResNet)}       & 61.01          & 73.33            & 62.44             & 58.10               & 63.72$\pm$6.7                                          \\ \cline{2-7} 
                                                            & CLIP \tiny{(ViT)}          & 64.25          & 68.54            & \textbf{66.29}             & 66.05               & 66.28$\pm$1.8                                          \\ \hline
    \textbf{Instagram}                                      & SWAG* \tiny{(RegNet)}      & 63.12          & \textbf{75.38}            & 62.96             & 64.61               & 66.52$\pm$6.0                                          \\ \hline
    \end{tabular}
\label{table:backbone-results}
\end{table}
\begin{table}[b]
    \scriptsize
    \centering
    \caption{Results of the binary classification task using different algorithms. The average accuracy and the standard deviation across the 4 domains in a single run are given, with the best accuracy (\%) highlighted in bold.}
    \begin{tabular}{l|cccc|c}
        \hline
        \textbf{Algorithm}       & \textbf{APTOS} & \textbf{EyePACS} & \textbf{Messidor} & \textbf{Messidor-2} & \textbf{Average Accuracy} \\ \hline
        \textbf{ERM }           & \textbf{95.42}          & 74.70            & \textbf{86.98}             & 77.47               & \textbf{83.63$\pm$9.5}                    \\
        \textbf{Fishr}         & 90.67          & 74.45            & 77.92             & \textbf{79.30}               & 80.59$\pm$7.0                     \\
        \textbf{DRGen}         & 82.05          & \textbf{75.41}            & 81.67             & 72.42                & 77.89$\pm$4.7                      \\
        \textbf{DGM-DR}        & 88.34          & 71.82            & 86.15             & 78.10               & 80.00$\pm$8.2                    \\
        \textbf{DGM-DR + SWAD} & 88.00          & 72.06            & 85.63             & 76.22               & 80.48$\pm$7.6                    \\ \hline
    \end{tabular}
\label{table:binary-classification}
\end{table}
\subsubsection{Changing the oracle pretraining datasets, methods, and backbones.}
We investigate the effect of changing the oracle on the DR classification task and report the results in Table \ref{table:backbone-results}. We use ImageNet pretrained ResNet-50 using Barlow Twins \cite{barlow-twins} and MoCo \cite{moco}, CLIP pretrained ResNet-50, and large-scale pretraining including CLIP pretrained ViT-B \cite{vit-b} and SWAG pretrained RegNetY-16GF \cite{regnet}. All experiments were performed with the same implementation details mentioned previously, except for RegNetY-16GF, where the batch size was changed from 32 to 16 due to hardware limitations.

\subsubsection{Binary classification of DR.}
We study the effect of changing the multiclass classification task into a binary classification task, where fundus images are classified as \emph{DR} or \emph{No DR}. The results of this experiment are reported in Table \ref{table:binary-classification}.

\section{Discussion}
In Table \ref{table:main-results}, we report the results of 4 different algorithms and show that DGM-DR outperforms all algorithms, including the previous state-of-the-art DRGen \cite{drgen} by a significant margin of 5.25\%. Additionally, DGM-DR demonstrates robustness with a relatively small standard deviation of 2.4 across three different experiments. As was concluded in \cite{domainbed}, ERM-based methods can outperform a range of DG methods, if carefully trained. We show in \ref{table:main-results} that the ERM method outperforms existing DG baselines that we compare with. On the other hand, we show that DGM-DR outperforms the ERM's performance for multiclass classification. We believe that even though the task of DR classification is challenging, the fundus images across all domains share common semantic structures, hence ERM is able to learn some domain-invariant features. However, the performance of DGM-DR is more stable, with a standard deviation being almost half that of ERM's. This can be attributed to DGM-DR's novel learning technique that aims to minimize a combination of cross entropy and mutual information regularization with an oracle, which enables it to learn more robust domain-invariant representations. Lastly, with the addition of SWAD to DGM-DR, performance further increases by a slight value (0.28\%), consistent with previous literature (e.g. \cite{rangwani2022closer}) where accuracy is improved when combined with SWAD.

In general, the performance of all algorithms on each of the datasets is consistent. This indicates that any decline or increase in performance of a dataset can be attributed to the distribution of the dataset itself, which is used as the target domain in the evaluation, and to the distribution of the combined source datasets on which the model is trained. For example, EyePACS \cite{EyePACS} consistently performs better across all algorithms. A possible yet arguable hypothesis is it is highly imbalanced, as demonstrated in Table \ref{table:datasets}, with the majority of images belonging to \emph{No DR}. Since the \emph{No DR} class is the majority in all datasets, the model is also biased towards it. Hence the model could be correctly classifying the \emph{No DR} case and randomly guessing in the other four cases.

In Table \ref{table:backbone-results}, we study the effect of changing the oracle pretraining datasets, methods, and backbones. The large SWAG* pretrained RegNetY-16GF oracle yields the best accuracy in this experiment, second only to our ResNet-50 with ImageNet ERM pretraining, possibly due to the smaller batch size and the limit of number of steps set for a fair comparison. In general, we observe that a larger oracle model trained on a bigger, more diverse dataset is able to guide the target model towards more generalized feature representations. However, it will require longer training time to converge.

In Table \ref{table:binary-classification}, we notice that ERM is doing a better job at binary classification of DR than DGM-DR. Since the binary classification problem is simpler, as is visually evident in Figure \ref{Figure:DR-images}, DG algorithms tend to negatively impact the results as they are likely to introduce more complexity. Furthermore, the generalization gap \cite{NEURIPS2021_ecf9902e} is typically smaller in binary classification than a multiclass setup. Therefore, ERM-based methods are likely to outperform DG-based methods in such scenarios.

The selection of the mutual information regularization coefficient $\lambda$, which controls the balance between the cross entropy loss and the mutual information regularization loss, is related to how informative the oracle model's knowledge is for the target model's task. A large $\lambda$ encourages the model to reduce the variance in the target features and enforce similarity between the target and oracle features. Thus, the model will focus on learning domain-invariant patterns originating from the oracle's knowledge, which is ImageNet in our main experiment. On the other hand, a small $\lambda$ reduces the emphasis on domain-invariance and thus may potentially lead to overfitting.

In our case, as shown in \cite{imagenet-init}, ImageNet initialization of deep learning models is beneficial in the context of medical imaging analysis, including fundus images. Therefore, we conclude that the best $\lambda$ for the case of DR classification is 1.0 for the ImageNet pretrained ResNet-50, in contrast with that of natural images where $\lambda$ is typically set to be $\{0.001, 0.01, 0.1\}$ as in \cite{MIRO}. We believe that in the DR classification problem, the oracle model has a significant impact on training the target DG model due to its rich low level feature representations which cannot be easily learnt from scratch or from a small size dataset.

As a final note, a very important part of a domain generalization solution is the model selection method, as it simplifies fair assessments by disregarding differences in results due to inconsistent hyperparameter tuning that may be attributed to the algorithms under study \cite{domainbed}. Furthermore, utilizing the test-domain validation set as a model selection method is inappropriate for a DG algorithm, which was done by DRGen \cite{drgen} in DR classification. Hence, one goal of this paper is to set a clear benchmark for DR classification using training-domain validation, thus allowing easy comparison with future work.

\section{Conclusion}
In this paper, we introduce DGM-DR to tackle the problem of DR classification with domain generalization. Our use of a large pretrained model to guide the target model towards learning domain-invariant features across different DR datasets through mutual information regularization achieves superior performance over the previous-state-of-the art DG methods. We also establish a clear benchmark for the task using a DG-appropriate model selection algorithm, thus allowing future work to make comparisons with our work.  
Further investigation to understand when and why DG-based methods could be superior or inferior to ERM-based approaches in medical imaging is needed. Although we believe that our work pushes the horizons of the DG field in medical image analysis, several DG-related research questions are yet to be investigated e.g., unsupervised DG, interpretable DG, and performance evaluation to DG methods.
%
%

\bibliographystyle{splncs04}
\bibliography{references.bib}

\newpage
\appendix

\center \textbf{Supplementary Material}

\begin{table}[]
\caption{Distribution of classes of each domain.}
\begin{tabular}{l|c|c|c|c|c|c}
\hline
\textbf{Dataset}    & \textbf{No DR} & \textbf{Mild} & \textbf{Moderate} & \textbf{Severe} & \textbf{Proliferative} & \textbf{Total \# of Images} \\ \hline
\textbf{EyePACS}    & 73.67\%        & 7.00\%        & 14.83\%           & 2.35\%          & 2.15\%                 & 88702                       \\
\textbf{APTOS}      & 49.25\%        & 10.12\%       & 27.29\%           & 5.28\%          & 8.06\%                 & 3657                        \\
\textbf{Messidor}   & 45.50\%        & 12.75\%       & 20.58\%           & 21.17\%         & 0.00\%                 & 1200                        \\
\textbf{Messidor-2} & 58.31\%        & 15.48\%       & 19.90\%           & 4.30\%          & 2.01\%                 & 1744                        \\ \hline
\end{tabular}
\label{table:datasets}
\end{table}

\begin{table}[h]
\caption{Results of changing the regularization coefficient $\lambda$ of DGM-DR. The average accuracy and the standard deviation across the 4 domains in a single run are given, with the best accuracy (\%) highlighted in bold.}
\begin{tabular}{l|cccc|c}
\hline
\textbf{Lambda Value} & \textbf{APTOS} & \textbf{EyePACS} & \textbf{Messidor} & \textbf{Messidor-2} & \textbf{Average Accuracy} \\ \hline
\textbf{0.001}        & 60.96         & 69.21           & 62.71            & 60.96               & 63.46$\pm$3.9                       \\
\textbf{0.01}         & 59.69         & 68.85           & \textbf{65.83}            & 65.97              & 65.09$\pm$3.9                    \\
\textbf{0.1}          & 60.92         & \textbf{73.61}           & 65.63            & 68.63              & 67.20$\pm$5.3                    \\
\textbf{1.0}          & \textbf{65.39}         & 70.12           & 65.63            & \textbf{69.41}              & \textbf{67.64$\pm$2.5}                    \\ \hline
\end{tabular}
\label{table:lambda-table}
\end{table}

\begin{figure}[h]
\centering
\begin{tabular}{cccc}
    \includegraphics[width=0.2\textwidth]{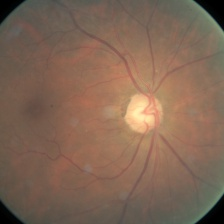} & \includegraphics[width=0.2\textwidth]{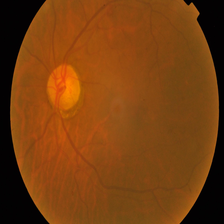} & \includegraphics[width=0.2\textwidth]{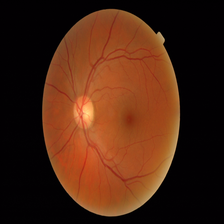} & \includegraphics[width=0.2\textwidth]{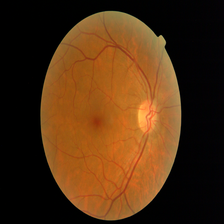} \\
    (a) EyePACs & (b) Aptos & (c) Messidor & (d) Messidor 2 \\
\end{tabular}
\caption{Sample images from the datasets, Grade 0}
\label{fig:g0}
\end{figure}

\begin{figure}[h]
\centering
\begin{tabular}{cccc}
    \includegraphics[width=0.2\textwidth]{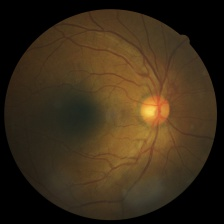} & \includegraphics[width=0.2\textwidth]{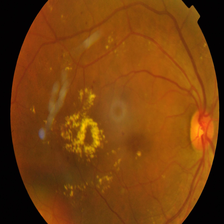} & \includegraphics[width=0.2\textwidth]{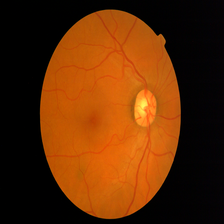} & \includegraphics[width=0.2\textwidth]{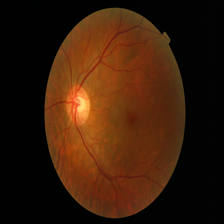} \\
    (a) EyePACs & (b) Aptos & (c) Messidor & (d) Messidor 2 \\
\end{tabular}
\caption{Sample images from the datasets, Grade 1}
\label{fig:g1}
\end{figure}

\begin{figure}[h]
\centering
\begin{tabular}{cccc}
    \includegraphics[width=0.2\textwidth]{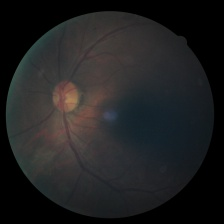} & \includegraphics[width=0.2\textwidth]{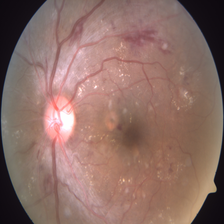} & \includegraphics[width=0.2\textwidth]{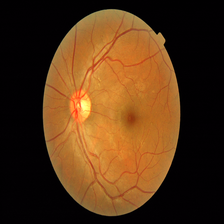} & \includegraphics[width=0.2\textwidth]{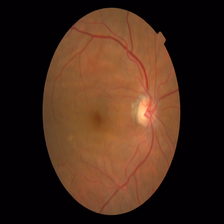} \\
    (a) EyePACs & (b) Aptos & (c) Messidor & (d) Messidor 2 \\
\end{tabular}
\caption{Sample images from the datasets, Grade 2}
\label{fig:g2}
\end{figure}

\begin{figure}[h]
\centering
\begin{tabular}{cccc}
    \includegraphics[width=0.2\textwidth]{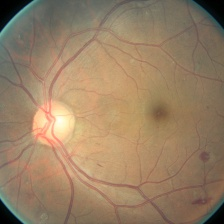} & \includegraphics[width=0.2\textwidth]{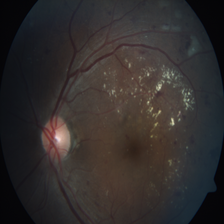} & \includegraphics[width=0.2\textwidth]{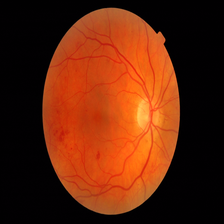} & \includegraphics[width=0.2\textwidth]{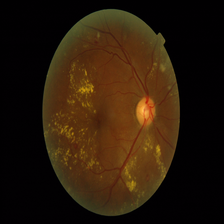} \\
    (a) EyePACs & (b) Aptos & (c) Messidor & (d) Messidor 2 \\
\end{tabular}
\caption{Sample images from the datasets, Grade 3}
\label{fig:g3}
\end{figure}

\begin{figure}[h]
\centering
\begin{tabular}{ccc}
    \includegraphics[width=0.2\textwidth]{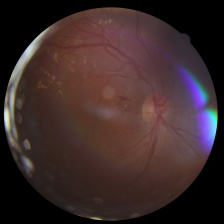} & \includegraphics[width=0.2\textwidth]{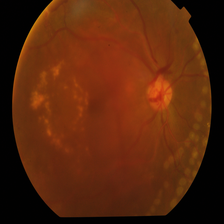} & \includegraphics[width=0.2\textwidth]{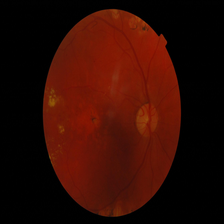}  \\
    (a) EyePACs & (b) Aptos & (c) Messidor 2 \\
\end{tabular}
\caption{Sample images from the datasets, Grade 4}
\label{fig:g4}
\end{figure}

\end{document}